\documentclass[conference, 10pt]{IEEEtran}

\usepackage{cite}
\usepackage{url}
\usepackage{graphicx}
\usepackage{color}
\usepackage{placeins}
\usepackage{float}
\usepackage{tabularx,colortbl}
\usepackage{ifthen}
\usepackage{soul}
\usepackage{tikz}
\newcommand{\bottomcopyrightbox}{%
\begin{tikzpicture}[remember picture,overlay]
\node[fill=white, inner sep=2pt]  
  at (current page.south) [yshift=1.5cm] 
  {978-1-7334677-3-5 © 2026 ACES};
\end{tikzpicture}%
}

\makeatletter

\newcounter{author}
\renewcommand{\author}[2][]{
   \stepcounter{author}
   \@namedef{author@\theauthor}{#2}
   \@namedef{authorlabel@\theauthor}{#1}
}

\newcounter{address}
\newcommand{\address}[2][]{
   \stepcounter{address}
   \@namedef{address@\theaddress}{#2}
   \@namedef{addresslabel@\theaddress}{#1}
}

\newcommand{\alsep}{and}

\def\newmaketitle{\par%
  \begingroup%
  \normalfont%
  \def\thefootnote{}
  \def\footnotemark{}
  \let\@makefnmark\relax
  \footnotesize
  \footnotesep 0.7\baselineskip
  \normalsize%
  \twocolumn[\thenewmaketitle\@IEEEaftertitletext]%
  \if@IEEEusingpubid
     \enlargethispage{-\@IEEEpubidpullup}%
  \fi
  \endgroup
  \setcounter{footnote}{0}\let\maketitle\relax\let\@maketitle\relax
  \gdef\@thanks{}%
  \let\thanks\relax}

\def\thenewmaketitle{
  \newpage
  \begin{center}%
    \vskip0.2em{\Huge\@IEEEcompsoconly{\sffamily}\@IEEEcompsocconfonly{\normalfont\normalsize\vskip 2\@IEEEnormalsizeunitybaselineskip
   \bfseries\large}\@title\par}\vskip1.0em\par%
    \vspace{1ex}
    \newcounter{c@author}
    \newcounter{c@tmp}
    \ifthenelse{\value{author}=2}{%
      \newcommand{\liand}{ and }}{%
      \newcommand{\liand}{, and }}
    \ifthenelse{\value{address}<2}{%
      \@nameuse{author@1}%
      \stepcounter{c@author}%
      \whiledo{\value{c@author}<\value{author}}{%
        \setcounter{c@tmp}{\value{author}}%
        \addtocounter{c@tmp}{-\value{c@author}}%
        \ifthenelse{\value{c@tmp}=1}{%
          \renewcommand{\alsep}{\liand}}{\renewcommand{\alsep}{, }}%
        \stepcounter{c@author}\alsep \@nameuse{author@\thec@author}}\\%
    }
    {
      \@nameuse{author@1}${}^{(\ref{\@nameuse{authorlabel@1}})}$%
      \stepcounter{c@author}%
      \whiledo{\value{c@author}<\value{author}}{%
      \setcounter{c@tmp}{\value{author}}%
      \addtocounter{c@tmp}{-\value{c@author}}%
      \ifthenelse{\value{c@tmp}=1}{%
        \renewcommand{\alsep}{\liand}}{\renewcommand{\alsep}{, }}%
      \stepcounter{c@author}\alsep \@nameuse{author@\thec@author}%
        ${}^{(\ref{\@nameuse{authorlabel@\thec@author}})}$%
      }
    }
    \vspace{0.2ex}

    \ifthenelse{\value{address}>0}{%
      \ifthenelse{\value{address}=1}{
        {\@nameuse{address@1}}
      }
      {
        \newcounter{c@address}

        \begin{center}
        \whiledo{\value{c@address}<\value{address}}
        {
          \refstepcounter{c@address}
            ${}^{(\thec@address)}$\,%
              \label{\@nameuse{addresslabel@\thec@address}}%
              \@nameuse{address@\thec@address}\\ %
        }
        \end{center}
      } 
    }
    {
      \relax
    }
  \end{center}
}

\makeatother

\title{Federated Learning-driven Beam Management in LEO 6G Non-Terrestrial Networks}

\author[org1]{Maria Lamprini Bartsioka}
\author[org1]{Ioannis A. Bartsiokas}
\author[org1]{Athanasios D. Panagopoulos}
\author[org1]{Dimitra I. Kaklamani}
\author[org1]{Iakovos S. Venieris}

\begin{document}
\newmaketitle
\bottomcopyrightbox

\begin{abstract}
Low Earth Orbit (LEO) Non-Terrestrial Networks (NTNs) require efficient beam management under dynamic propagation conditions. This work investigates Federated Learning (FL)-based beam selection in LEO satellite constellations, where orbital planes operate as distributed learners through the utilization of High-Altitude Platform Stations (HAPS). Two models, a Multi-Layer Perceptron (MLP) and a Graph Neural Network (GNN), are evaluated using realistic channel and beamforming data. Results demonstrate that GNN surpasses MLP in beam prediction accuracy and stability, particularly at low elevation angles, enabling lightweight and intelligent beam management for future NTN deployments.
\end{abstract}

\section{Introduction}
Sixth-generation (6G) systems aim to support the increasing need for global and real-time connectivity, as well as hyper-reliable and low-latency communications, providing a platform for new use case scenarios across different industries \cite{11018294}. Existing terrestrial networks cannot always meet the aforementioned requirements on their own. Thus, Non-Terrestrial Networks (NTNs) are proposed within 6G to formulate an integrated global network that serves seamlessly heterogeneous communication scenarios. NTNs comprise three tiers. The aerial and satellite tiers handle resource management, coordination, and reliable connectivity by utilizing Unmanned Aerial Vehicles (UAVs), High-Altitude Platform Stations (HAPS) and geostationary, medium, and low earth orbits (GEO, MEO, LEO) satellites; while the terrestrial tier supports edge operations with ground base stations and users \cite{9846947}.

To satisfy the stringent throughput requirements of emerging 6G services, systems increasingly rely on large-scale Multiple-Input Multiple-Output (MIMO) techniques \cite{5473886}. In this context, beamforming plays a critical role by directing transmission power toward specific spatial directions, compensating for severe path loss and enabling reliable high signal-to-noise-ratio (SNR) links. Beam management is, therefore, essential for maintaining robust connectivity under high-mobility and dynamic channel conditions. However, conventional beam selection approaches-such as CSI-based estimation- suffer from computational complexity, signaling overhead, and limited adaptability. This makes them sub-optimal for dynamic NTN deployments. \cite{marenco2024machine}.

Federated learning (FL) provides an efficient and privacy-preserving solution for beam selection in NTNs by enabling distributed model training without raw data exchange \cite{10423251}. This is particularly critical in LEO environments, where centralized learning incurs excessive signaling overhead and latency. By leveraging the multi-tier NTN architecture for decentralized model aggregation, FL enables scalable and adaptive beam management under mobility and intermittent connectivity.

This work focuses on LEO mega-constellations and proposes a novel FL beam management framework leveraging geometrical information to enable accurate, robust, and scalable beam prediction that maximizes the system's SNR.

\section{System Model and FL Framework}

\subsection{Satellite Scenario}
\begin{figure}[h!]
\begin{center}
\noindent
  \includegraphics[width=2.9in]{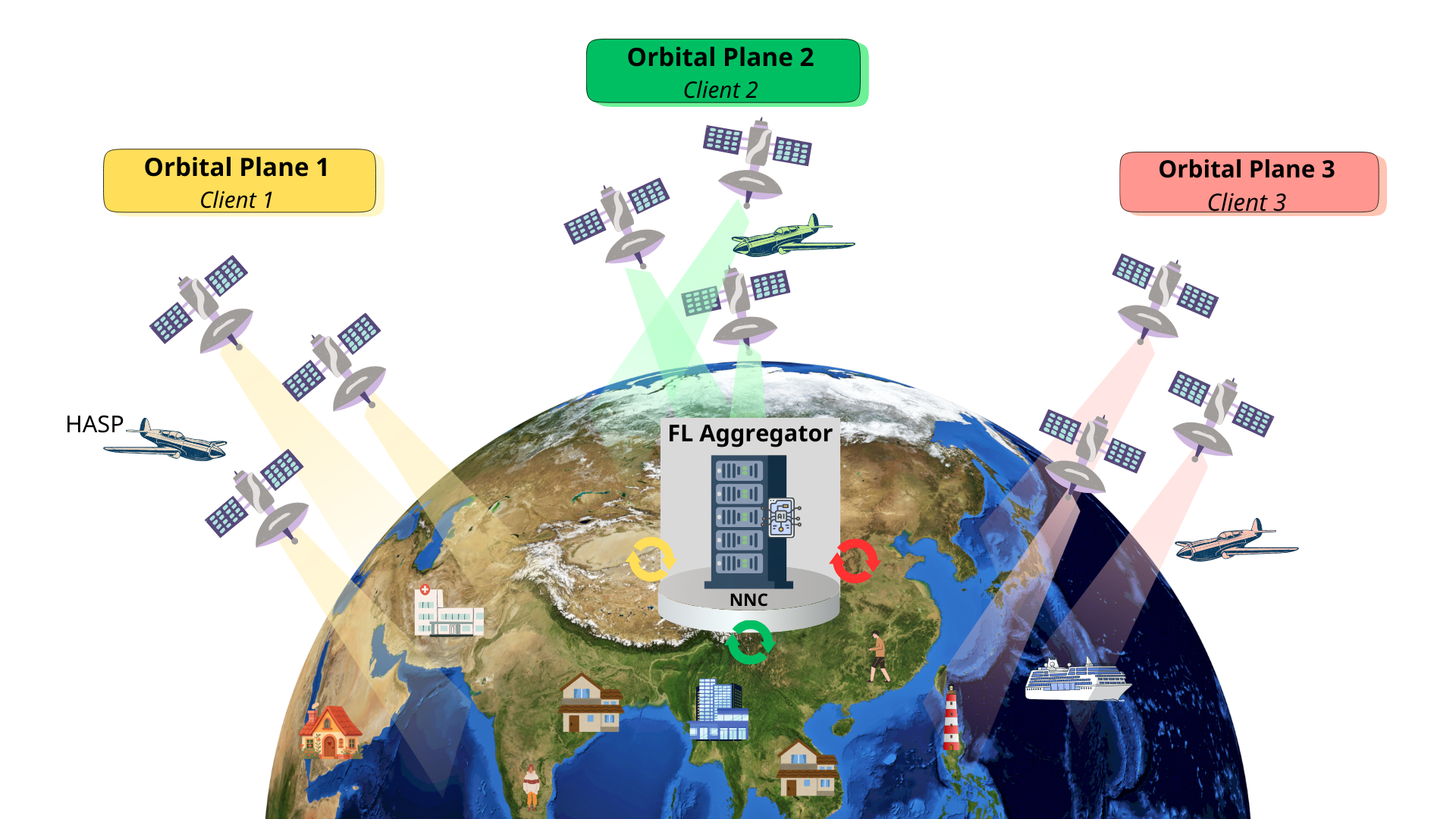}
  \caption{NTN topology with FL support}\label{fig1}
\end{center}
\end{figure}

We consider a LEO constellation as part of an NTN (Fig.~\ref{fig1}), operating at the S-band (downlink frequency 2 GHz) and a range of altitude between 1015-1325 km. The constellation consists of multiple orbital planes with their satellites acting as transmitters, and multiple UEs acting as receivers \cite{matlab}. Let $\mathcal{P} = \{1, 2, \dots, P\}$ denote the set of orbital planes and $\mathcal{S_P} = \{1, 2, \dots, S\}$ the set of uniformly distributed satellites moving along orbital plane $p \in \mathcal{P}$. A set of $\mathcal{U} = \{1, 2, \dots, U\}$ static UEs is also deployed in a predefined Region of Interest (ROI). UEs are equipped with small Uniform Planar Array (UPA) receiver antennas and are subject to a minimum elevation angle constraint $\theta_{min}$. Similarly, each satellite is equipped with a 4×4 UPA antenna and supports a beam codebook corresponding to fixed steering directions in azimuth and elevation. 

The system is simulated for two hours, while a number of snapshots (1000) is selected to evaluate the links and gather the appropriate information that will form the input dataset for the Machine Learning (ML) problem. At each snapshot $t \in \{1, 2, \dots, T\}$, the relative satellite–UE geometry is assumed quasi-static, while the links are filtered in only if $\theta_{u,s}(t) > \theta_{min}$, where $\theta_{u,s}(t)$ stands for the elevation angle between UE $u \in \mathcal{U}$ and satellite $s \in \mathcal{S_P}$. For each visible link at snapshot $t$ the satellite evaluates all candidate beams. Consequently, the dataset produced includes scenario information, geographic features, satellite/beam azimuth and elevation angles and other geometrical characteristics.

From the exhaustive search of the simulation, the supervised learning target is also computed as:  
\begin{equation}
b_{u,s}^\star(t)=\arg\max_{b} SNR_{u,s,b}
\end{equation}
where the $SNR$ takes into account transmit array gain, path loss, small-scale elevation-dependent Rician fading, temporally correlated shadowing, and receiver noise power \cite{espineira2008modeling}.

\subsection{Federated Learning Scheme}
Satellite networks are characterized by distributed heterogeneity across orbital planes. For this reason, we propose a hierarchical FL framework where each orbital plane constitutes a logical FL client. Satellites belonging to the same plane coordinate through an intra-plane level HAPS, which acts as edge aggregator and control entity. There local data are consumed to train a plane-level local model. After a fixed number of local training epochs (200), model parameters are transmitted to a ground-based centralized Network Control Center (NNC), which performs inter-plane federated averaging (FedAvg) to obtain a global model. The aggregated global model is then broadcasted back to all clients for the next training round. In this way, we ensure data locality, reduce communication overhead and capture non-IID data distributions induced by distinct orbital geometries, all under realistic NTN constraints.


Beam selection is addressed using two learning architectures: a Multi-Layer Perceptron (MLP) and a Graph Neural Network (GNN), both predicting the beam that maximizes the received SNR for a given UE–satellite snapshot. The MLP consists of fully connected layers with non-linear activations and evaluates each candidate beam independently, offering a lightweight and computationally efficient solution suited to geometry-driven scenarios. In contrast, the GNN models beams as graph nodes connected to neighboring beams and employs graph convolutions to exchange information across beams, enabling the learning of relative beam quality and local consistency.

\section{Performance Evaluation}
After data collection, ML models are trained under the FL setup and evaluated using the same test-set. Table~\ref{table1} reports the overall performance comparison. The GNN significantly outperforms the MLP in all metrics, showing a higher probability of predicting the optimal beam at each snapshot, demonstrating improved robustness under the non-IID data distributions. Both models achieve near-perfect Top-3 accuracy, confirming that the optimal beam is almost always included among the top candidates. However, there is a trade-off between better performance and increased model complexity. Model size and training time remain small, though, preserving practical feasibility for NTN deployment.

\begin{table}[t]
\begin{center}
\caption{Beam Prediction Performance Comparison under FL Setup.} \label{table1}
\begin{tabular}{|c|c|c|}
 \hline
 Metric & MLP & GNN\\
 \hline
 Top-1 Accuracy & 88.41\% & 96.14\%\\
 \hline
 Top-3 Accuracy & 98.03\% & 99.52\%\\
 \hline
 Top-1 Mean Accuracy across clients & 89.15\% & 95.97\%\\
 \hline
 Training Time & 6.27 sec & 19.23 sec\\
 \hline
 Model Size & 38.40 KB & 71.43 KB\\
 \hline
\end{tabular}
\end{center}
\end{table}

\begin{figure}[h]
\centering
\begin{minipage}[t]{0.25\textwidth}
\centering
\includegraphics[width=\textwidth,height=3cm]{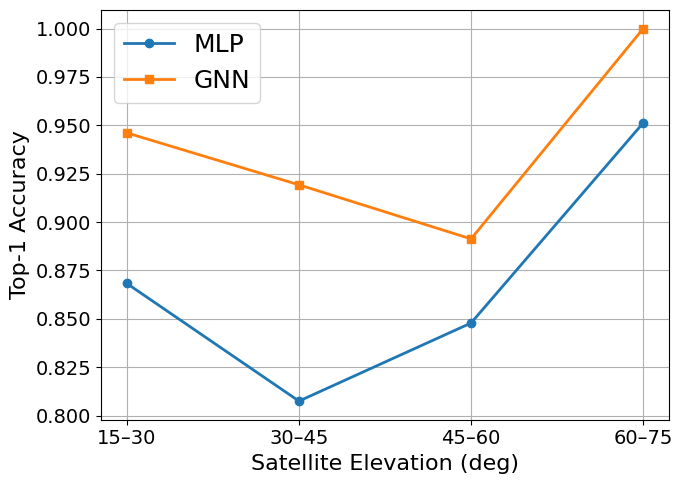} 
\label{fig2:subfig1}
\end{minipage}%
\begin{minipage}[t]{0.25\textwidth}
\centering
\includegraphics[width=\textwidth,height=3cm]{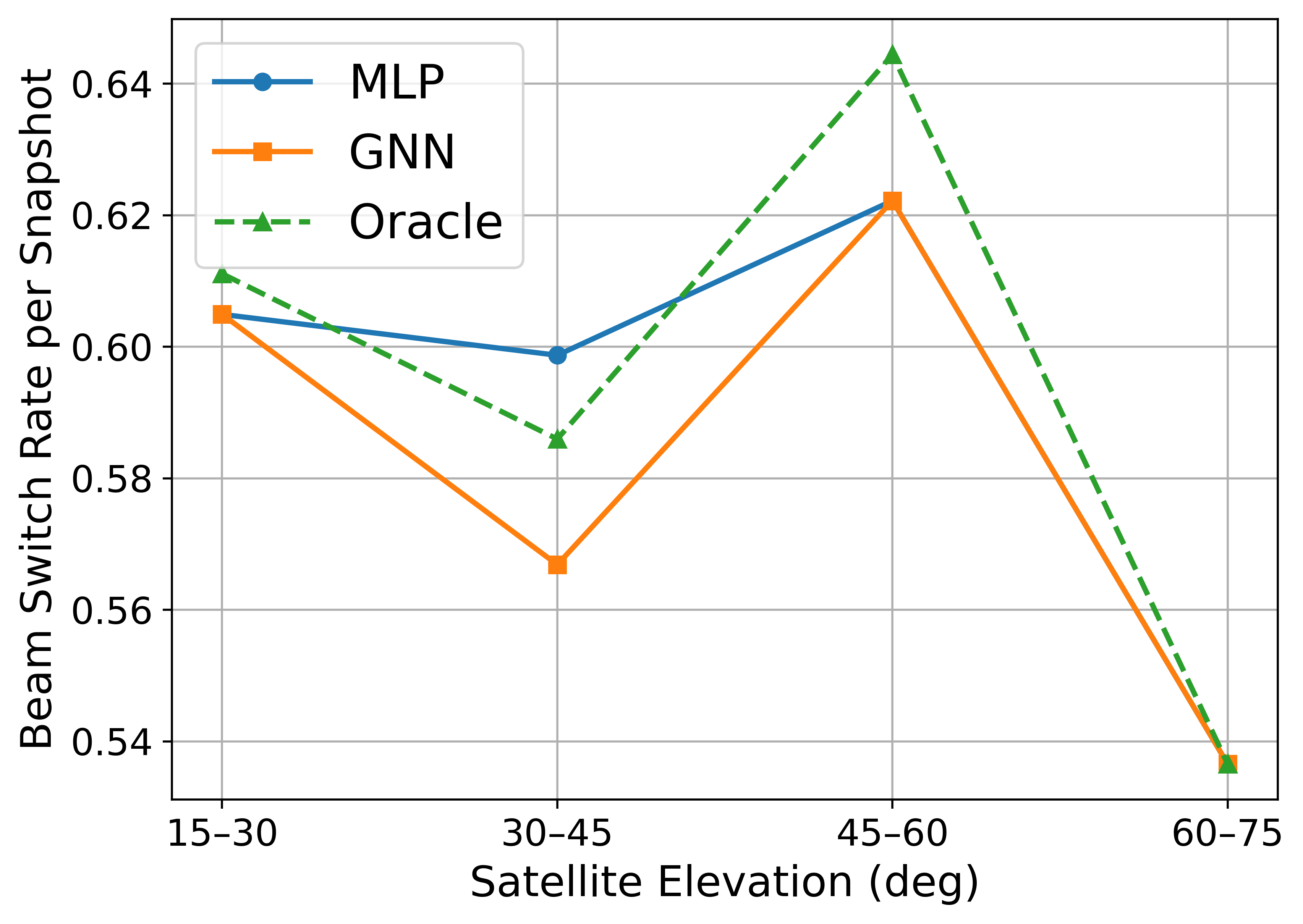}
\label{fig2:subfig2}
\end{minipage}
\caption{MLP and GNN performance under different satellite elevation angle (a) global top-1 accuracy (b) beam switching.}
\label{fig:fig2}
\end{figure}

To further examine system-level behavior, Fig.~\ref{fig:fig2} illustrates the impact of satellite elevation on beam selection accuracy and stability. As expected, accuracy improves with increasing elevation for both models due to more favorable propagation conditions, with GNN leading the race. Similarly, all approaches exhibit higher switching rates at low elevations due to increased channel dynamics, fading and shadowing. Importantly, the GNN closely follows the oracle switching trend, derived from maximum-SNR beam in the simulation, while the MLP exhibits slightly higher switching at low elevation. Overall, the results demonstrate that the GNN maintains beam stability consistent with the underlying channel characteristics.

\section{Conclusions and Future Extensions}
This work presented a Federated Learning framework for beam prediction in LEO NTNs, demonstrating that GNN outperforms MLP by effectively capturing inter-beam relationships, achieving 8.7\% higher accuracy and 1.8\% more stable beam switching with limited complexity overhead. Future work will extend this approach toward fully hierarchical FL architectures and interference-aware beam management with joint user–satellite optimization.



%

\bibliographystyle{IEEEtran}
\bibliography{mybibfile}

\end{document}